\documentclass{vgtc}

\usepackage{amsmath,amssymb}
\vgtccategory{Research}
\vgtcinsertpkg

\usepackage{times}

\usepackage{mathptmx}
\usepackage{booktabs}
\usepackage{graphicx}
\usepackage{xcolor}
\usepackage[most]{tcolorbox}
\usepackage{balance}

\graphicspath{{figures/}{figures/circuit_plots_v2/}{figures/plots_v2/}{pictures/}{images/}{./}}

\newcommand{\stage}[1]{\texttt{#1}}
\newcommand{\safeincludegraphics}[2][]{%
\IfFileExists{#2}{\includegraphics[#1]{#2}}{%
\fbox{\parbox[c][0.75in][c]{0.92\linewidth}{\centering Missing figure: \texttt{\detokenize{#2}}}}}%
}

\definecolor{stageblue}{HTML}{1F5FA8}
\definecolor{stageblueback}{HTML}{EEF6FF}
\newtcolorbox{contribbox}{enhanced,colback=stageblueback,colframe=stageblue,boxrule=0.7pt,arc=1.5mm,left=4pt,right=4pt,top=4pt,bottom=4pt,before skip=0.5em,after skip=0.5em}

\definecolor{stageblue}{HTML}{1F5FA8}
\definecolor{stageblueback}{HTML}{EEF6FF}

\newtcolorbox{traceexamplebox}{enhanced,
  colback=stageblueback,
  colframe=stageblue,
  boxrule=0.7pt,
  arc=1.5mm,
  left=4pt,right=4pt,top=4pt,bottom=4pt,
  before skip=0.55em,after skip=0.55em,
  fonttitle=\small\bfseries}

\usepackage{orcidlink}
\newcommand{\affmark}[1]{\textsuperscript{#1}}

\title{Visualizing Graph-to-Answer Mechanism Recovery in Materials-Science Hypothesis Generation}

\author{%
\parbox{0.98\textwidth}{\centering
Shashwat Sourav\affmark{1,2,3,4}\thanks{e-mail: s.shashwat@wustl.edu}
\quad
Subhadeep Pal\affmark{5}\thanks{e-mail: spu8516@mit.edu}
\quad
Markus J. Buehler\affmark{5,6,7}
\\[-0.15em]
Sanjay Das\affmark{2}
\quad
Fiona Y. Wang\affmark{8}
\quad
Dominik So\'os\affmark{9,2}
\quad
Tirthankar Ghosal\affmark{2}
\\[0.45em]
{\scriptsize
\begin{tabular}{@{}c@{}}
\affmark{1}Department of Physics, Washington University in St. Louis \quad
\affmark{2}Oak Ridge National Laboratory \quad
\affmark{3}Lawrence Berkeley National Laboratory
\\[-0.1em]
\affmark{4}UniverseTBD \quad
\affmark{5}Department of Civil and Environmental Engineering, Massachusetts Institute of Technology
\\[-0.1em]
\affmark{6}Department of Mechanical Engineering, Massachusetts Institute of Technology
\\[-0.1em]
\affmark{7}Schwarzman College of Computing, Massachusetts Institute of Technology 
\\[-0.1em]
\affmark{8}Department of Biological Engineering, Massachusetts Institute of Technology
\quad \affmark{9}Department of Computer Science, Old Dominion University
\end{tabular}
}
}
}

\abstract{AI co-scientists can generate fluent materials-science hypotheses, but fluency does not show that an answer preserves a scientifically meaningful mechanism. We present a graph-to-answer mechanism-tracing case study for Graph-PRefLexOR-8B, a Qwen3-8B model adapted to expose distinct stages for brainstorming, graph construction, pattern extraction, and synthesis. We organize semantic backtracking, graph corruption, activation-based recovery measurements, and layer-by-token-region grids into a visual diagnostic workflow for inspecting this pathway. Across 100 open-ended materials-science questions, final answers remain closest to the model's own structured stages, especially synthesis. Under graph corruption, a full sweep over 37 residual-stream checkpoints---the embedding output and 36 transformer blocks---shows little mechanism recovery in the earlier transition region at layers 7--10, recovery instead concentrates in late synthesis and answer-start regions around layers 30 and 36. The workflow is intended to help scientists and model developers identify where a generated hypothesis loses or regains mechanism support before it is passed to downstream experimental planning.}


\begin{document}
\raggedbottom
\firstsection{Introduction}
\maketitle

A useful materials-science hypothesis is not just a plausible sentence. It usually has a causal form: a problem or bottleneck motivates an intervention, the intervention changes a mechanism, the mechanism affects a target property, and the property leads to an expected outcome~\cite{agrawal2016materials,2016MRSBu..41..587M,2025arXiv250113299K,2025AcMat.29721307L}. This structure matters because generated scientific answers can sound fluent while leaving the central mechanism unsupported. A model may mention an interface, a polymer additive, a crosslinking process, or an improved toughness target but still fail to show which relation makes the proposed intervention scientifically plausible.

We treat this as a visual model-auditing task. Here, ``visual'' refers to linked static views of the reasoning trace, corruption effects, and the full layer-by-token-region recovery landscape. These views support analytical comparison, but the current contribution is not an interactive visual-analytics interface or a user study. Existing language models can produce long chain-of-thought-style explanations, yet the visible trace is not guaranteed to be faithful to the final answer~\cite{wei2022chain,2023arXiv230713702L,2023arXiv230504388T,2024arXiv240214897B}. Knowledge graphs and graph-augmented generation make entities and relations more explicit~\cite{2023arXiv230608302P,wang2023survey}, but they are often treated as external context rather than as an intermediate object that can be perturbed and compared with downstream model behavior. This leaves a gap between the readable trace shown to a scientist and the internal representations associated with the final answer.

\begin{figure*}[htbp]
    \centering
    \includegraphics[width=\linewidth]{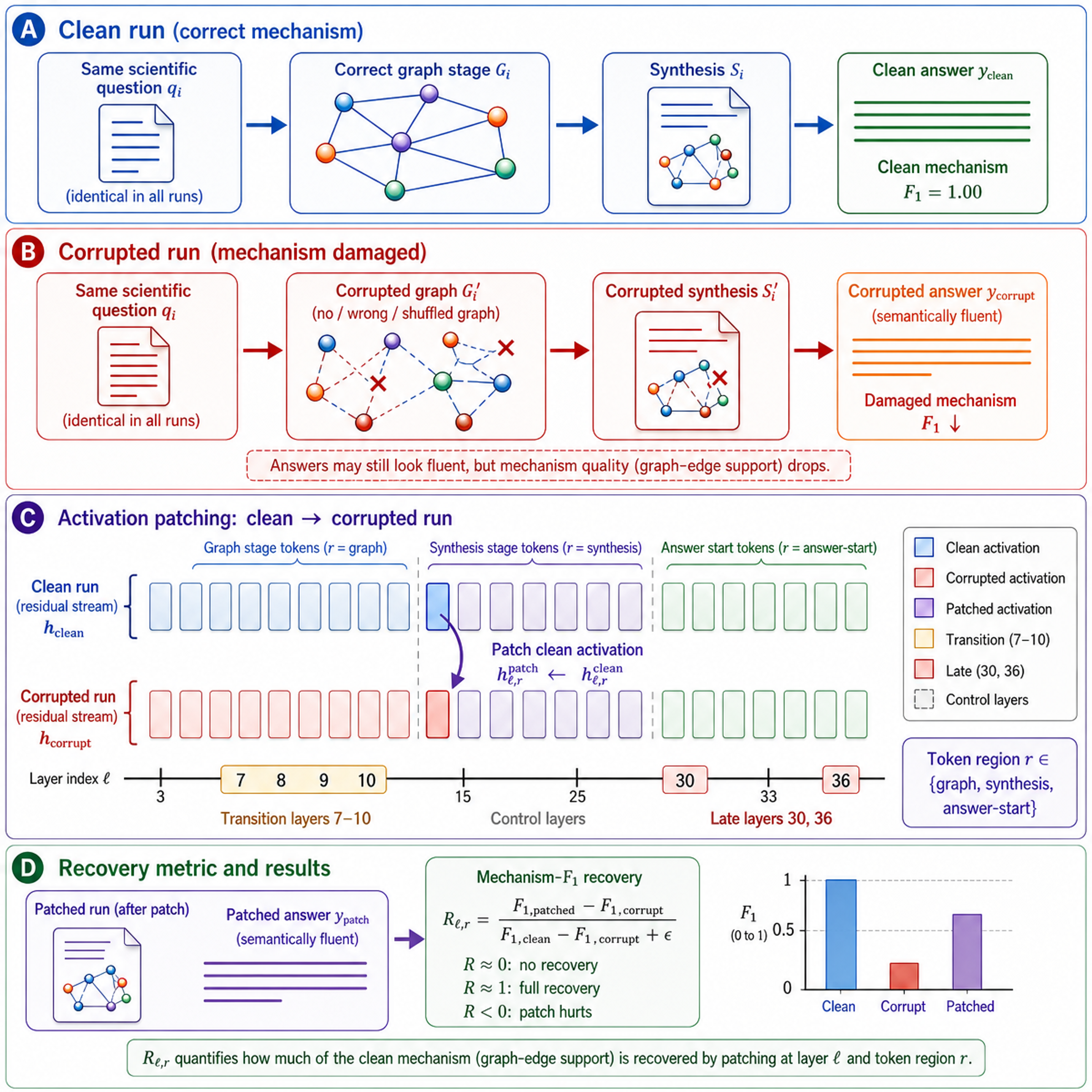}
     \caption{Graph-to-answer mechanism-tracing workflow. The linked views compare a clean structured trace with graph-corrupted runs and summarize where mechanism information is recoverable across layers and token regions. The current paper uses these as static diagnostic views rather than as an interactive interface.}
    \label{fig:placeholder}
\end{figure*}

We study this gap using Graph-PRefLexOR-8B, an 8B graph-native reasoning model initialized from Qwen3-8B and adapted to produce explicit, sentinel-delimited scientific reasoning stages~\cite{2026arXiv260700924P}. Qwen3-8B later appears as the matched backbone baseline, it is not a second version of Graph-PRefLexOR. The model produces the following staged trace:
\begin{equation}
\begin{split}
\stage{<brainstorm>} \rightarrow \stage{<graph>} \rightarrow \stage{<graph\_json>}\\
\rightarrow \stage{<patterns>}  \rightarrow \stage{<synthesis>} \rightarrow \text{answer}
\end{split}
\end{equation}
This staged output gives us visible objects to compare and do perturbations. The main question we are investigating is: \emph{where does a graph-structured scientific mechanism become recoverable inside the model's answer pathway?}

\begin{contribbox}
\normalsize
\textbf{Contribution:} 
We introduce a visual diagnostic workflow for graph-native scientific reasoning models. It links three questions: (i) does the final answer remain semantically tied to the visible reasoning stages, (ii) does corrupting the graph damage mechanism content even when the answer remains fluent, and (iii) where does the mechanism-recovery signal concentrate across layers and token regions?
\end{contribbox}

\paragraph{Intended use in automated science.}
The intended users are scientists supervising an AI hypothesis generator and model developers diagnosing it. The workflow provides a human checkpoint before a generated hypothesis is promoted to literature search, simulation, or experiment selection. A user can compare corruption conditions, locate whether recovery is diffuse or concentrated, and then inspect the corresponding graph and synthesis text. A hypothesis with a fluent answer but weak mechanism support can be flagged for regeneration, abstention, or domain-expert review. The heatmap is therefore not only a presentation device: it helps determine which stage and layer region warrants closer investigation and prevents a conclusion from resting on a few selected layers.

Our analysis combines three levels of evidence. First, semantic backtracking~\cite{2024arXiv241007295U} asks whether the final answer is closest to the model's own reasoning trace or to another reference text. We use this only as a consistency check, not as causal evidence. Second, graph corruptions test whether removing, replacing, or shuffling the graph stage weakens the scientific mechanism in the final answer. Third, activation patching tests where clean internal states restore mechanism content in a corrupted run~\cite{2022arXiv220205262M,wang2022interpretability,zhang2023towards}. The key visualization is a full layer-by-token-region recovery grid over 37 residual-stream checkpoints (the embedding output plus 36 transformer blocks) and all main trace regions. This grid is important because it avoids selecting layers after the fact.

Across 100 materials-science questions, we observed that the full recovery grid shows layers 7-10 (which show a representational transition between reasoning and answering), do not recover the mechanism under patching. Mechanism recovery appears later, especially in synthesis and answer-start regions near layers 30 and 36. This suggests that graph reasoning is not simply copied into the answer, it is transformed into answer-ready mechanism content late in the network.

\section{Related Work and Positioning}

\textbf{Reasoning traces and faithfulness.}

Chain-of-thought prompting can improve reasoning performance~\cite{wei2022chain}, but visible reasoning should not be treated as a direct explanation of the model's internal computation. Prior work shows that chain-of-thought can be incomplete or unfaithful to the answer~\cite{2023arXiv230713702L,2023arXiv230504388T,2024arXiv240214897B}. In this work we take this information into account and use visible graph traces as inspectable hypotheses about the model's reasoning. Finally, we test them with perturbation and patching.

\noindent\textbf{Graph-structured scientific reasoning.}
Graph-based representations are natural for scientific explanation because they make entities, relations, and mechanisms explicit. Graph retrieval and graph-augmented generation have been used to improve factual grounding and relational reasoning~\cite{2023arXiv230608302P,wang2023survey,2026arXiv260700924P}. In our setting we have the graph as not only an input artifact, but an intermediate model-generated stage whose influence on the final answer can be visualized and tested.

\noindent\textbf{Mechanistic interpretability and activation patching.}
Mechanistic interpretability asks which internal model states are associated with a behavior rather than treating the model only as an input--output system. Activation patching provides one intervention-based test: run the model normally, deliberately corrupt part of the computation, restore a clean internal state at one location, and measure whether the behavior recovers~\cite{2022arXiv220205262M,wang2022interpretability,zhang2023towards}. The residual stream is the running vector representation passed from one transformer block to the next. Patching conclusions can change with the corruption, metric, and patch site, so we show the full layer-by-region landscape and compare it with matched controls rather than interpreting a single selected layer.

\noindent\textbf{Visual diagnostic contribution and scope.}
The visual contribution is a coordinated set of static diagnostic views: readable stage traces, corruption-strength summaries, and recovery heatmaps indexed by model layer and token region. The heatmaps expose patterns that aggregate scores hide, including whether recovery is broad or localized, whether the same location persists across corruptions, and whether a previously hypothesized transition region actually restores mechanism content. We use the term \emph{visual diagnostic workflow} rather than claim a complete interactive visual-analytics system. Building an interface that supports filtering, case retrieval, linked selection, and prospective evaluation with scientists is future work.

\section{Model, Data, and Analysis Task}

Graph-PRefLexOR-8B is initialized from Qwen3-8B and adapted to produce structured scientific reasoning traces~\cite{2026arXiv260700924P}. We use this model for three practical reasons. First, its sentinel-delimited \stage{<graph>}, \stage{<patterns>}, and \stage{<synthesis>} stages provide inspectable intermediate objects. Second, its released weights permit hidden-state analysis that is not possible with closed API models. Third, the 8B model is the strongest scale evaluated in the prior Graph-PRefLexOR study while remaining feasible for a full layer-by-region sweep. Qwen3-8B is used as the matched base-model comparison because it is the backbone from which Graph-PRefLexOR-8B was adapted and already supports an explicit thinking mode. This selection supports a controlled case study, it does not establish that Qwen is uniquely suitable or that the result transfers to Claude, GPT, or Llama models.

We reuse the public 100-question evaluation set introduced with Graph-PRefLexOR~\cite{2026arXiv260700924P}. The set is not a standard materials-science benchmark. It was curated from published materials science and mechanics literature to probe five forms of open-ended reasoning: cross-domain mapping, causal multiscale reasoning, hidden-variable identification, model abstraction and breakdown, and tradeoff or non-monotonicity reasoning. Reusing it allows the tracing analysis to be compared directly with the model's prior answer- and trace-level evaluation. The benchmark and training data are separate, but both the domain and question count limit the scope of our conclusions.

Each question is designed to require a mechanism path rather than a single factual answer. We represent the intended answer structure as
\[
p_i \rightarrow f_i \rightarrow u_i \rightarrow m_i \rightarrow t_i \rightarrow o_i,
\]
where $p_i$ is the problem context, $f_i$ the failure mode, $u_i$ the intervention, $m_i$ the mechanism, $t_i$ the target property, and $o_i$ the expected outcome.

\paragraph{Analysis requirements.}
The workflow is organized around three requirements.

\textbf{Requirement 1 (Coverage and comparison).}
The visual summary should expose the full layer-by-token-region recovery landscape rather than a few selected layers. Activation-patching conclusions can depend on the metric, corruption, and patch site~\cite{zhang2023towards}. We therefore sweep all residual-stream checkpoints and the main trace regions and summarize the result in \Cref{fig:full_sweep} and Table~\ref{tab:full_sweep_bootstrap}. A reader uses the grid in three passes: compare columns to identify the affected trace stage, compare rows to distinguish early transition from late recovery, and compare corruption panels to determine whether the pattern persists when graph content is absent or misleading.

\textbf{Requirement 2 (Mechanism specificity).}
The score should test whether the directed scientific mechanism is restored, not merely whether the patched answer remains fluent or close to the clean answer in embedding space. We therefore define mechanism-F1 over extracted directed relations (\Cref{sec:mechanism_f1}) and separately show that high answer similarity can coexist with mechanism damage (\Cref{sec:embedding_hides}).

\textbf{Requirement 3 (Matched controls).}
Recovery should be interpreted relative to controls because injecting a clean representation can alter output even when it is not mechanism-specific. We compare graph corruptions, non-target layers, random-region patching, and clean-run sensitivity checks (\Cref{sec:controls_validation}). These comparisons support a localized-recoverability claim under the stated intervention and metric, they do not by themselves identify a complete neuron- or head-level circuit.

\section{Semantic Backtracking as a Consistency Check}
\label{sec:semantic_backtracking}

We first ask a descriptive question: which visible text is closest in meaning to the final answer? We embed the final response and each candidate source with the same sentence-embedding model, normalize the vectors, and compute cosine similarity~\cite{2024arXiv240305440S}. The candidate with the highest similarity is called the semantic backtracking source.

For Qwen3-8B, the candidates are its own thinking trace, the Graph-PRefLexOR-8B final answer, and Graph-PRefLexOR-8B's \stage{<brainstorm>}, \stage{<graph>}, \stage{<patterns>}, and \stage{<synthesis>} stages. For Graph-PRefLexOR-8B, we perform the symmetric comparison using its own structured stages and the corresponding Qwen3-8B outputs. This is only a consistency check: semantic proximity does not establish that one text caused another.

The result is asymmetric. Qwen3-8B final answers are closest to their own thinking trace in 16 of 100 cases, whereas Graph-PRefLexOR-8B final answers are closest to one of their own structured stages in 92 of 100 cases, most often \stage{<synthesis>}. This motivates the intervention-based analysis below: when the graph is damaged, where does mechanism-related information become recoverable along the graph-to-synthesis-to-answer pathway?

\section{Mechanism Recovery Workflow}

\subsection{Graph corruptions}
\label{sec:graph_corruptions}

\begin{figure}[t]
\centering
\includegraphics[width=\linewidth]{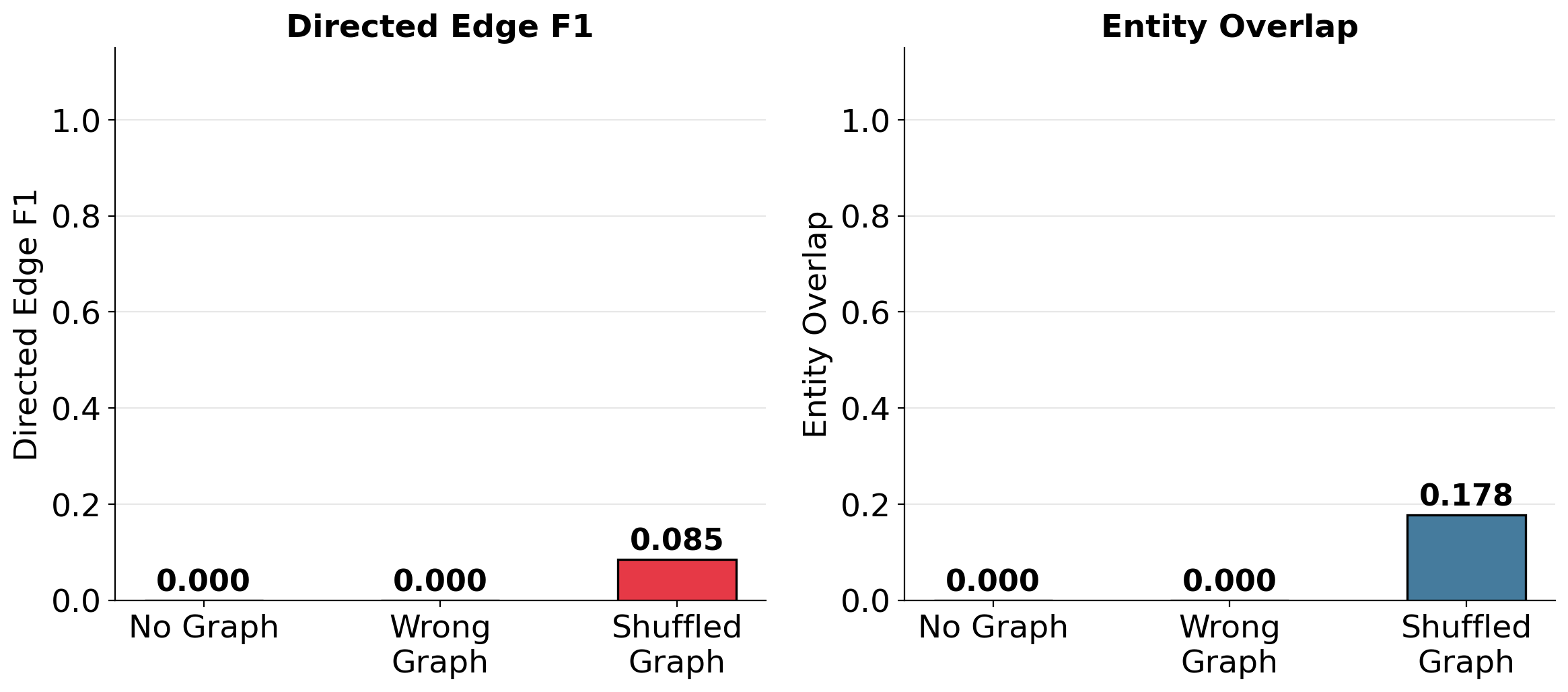}
\caption{Edge-level corruption strength. No-graph and wrong-graph corruptions remove the clean directed graph structure, while shuffled-graph preserves only small entity overlap and little directed-edge structure.}
\label{fig:edge_corruption}
\end{figure}

To analyze the dependency of the model on structural information, we proposed three variants of graph corruption alongwith the clean baseline native to the graph (Figure \ref{fig:edge_corruption}). These variants manipulate the reasoning trace as follows: (1) no-graph, where the graph stage is completely removed, (2) wrong-graph, where the graph is replaced by an unrelated question's graph, and (3) shuffled-graph, where edges are permuted to preserve a graph-like syntax while severing directed causal relations. Quantitative validation of these structural interventions, as depicted in Figure \ref{fig:edge_corruption} confirms significant topological degradation. Specifically, directed-edge F1 scores fall to 0.000 for both the no-graph and wrong-graph setups, and to 0.085 for the shuffled-graph condition, with the latter maintaining a baseline entity overlap of 0.178. These verified structural degradations establish a rigorous framework for evaluating the causal impact of graph-structured reasoning on final model generations.

\subsection{Mechanism-F1}
\label{sec:mechanism_f1}


To quantitatively track the survival and transmission of structural knowledge within the model, we require an explicit metric that operates over graph topologies rather than raw text surfaces. To achieve this, we introduce an edge-level extraction approach to map free-text generations back onto a structured, ground-truth reference space.

Let $E_i^{\mathrm{ref}}$ denote the reference mechanism path for question $i$, represented as a set of directed typed edges over the path fields $p,f,u,m,t,o$. For an answer $y$, we extract a generated edge set $E(y)$ after lower-casing, punctuation removal, stage-tag stripping, and simple lexical canonicalization. Directionality is preserved. We compute
\[
P(y)=\frac{|E(y)\cap E_i^{\mathrm{ref}}|}{|E(y)|+\epsilon}, \quad
Q(y)=\frac{|E(y)\cap E_i^{\mathrm{ref}}|}{|E_i^{\mathrm{ref}}|+\epsilon},
\]
and
\[
M(y)=\frac{2P(y)Q(y)}{P(y)+Q(y)+\epsilon}.
\]
Thus $M(y)$ measures edge-level agreement with a question-level mechanism reference, not merely similarity to the clean model answer. We use $\epsilon=10^{-8}$ for numerical stability. 



We report the manual validation of this extractor together with the robustness controls in \Cref{sec:controls_validation}. Thus \(M(y)\) measures edge-level agreement with a question-level mechanism reference, not merely similarity to the clean model answer.
We use \(\epsilon=10^{-8}\) for numerical stability. We report the manual validation of this extractor together with the robustness controls in \Cref{sec:controls_validation}.

\subsection{Activation patching}
\label{sec:activation_patching}


Activation patching compares a clean run with a deliberately corrupted run. At one selected layer and token region, the corrupted internal state is replaced with the corresponding clean state, the rest of the corrupted computation is left unchanged. If the downstream mechanism score improves, that location contains information that is recoverable under this intervention. This is a localized intervention test, not proof that the selected state is the model's only causal route.

For question $i$, let $y_i^{\mathrm{clean}}$ be the clean answer and $y_i^{\mathrm{corr}}$ the answer after a graph corruption. We patch clean residual-stream activations into the corrupted run at layer $\ell$ and token region $r$:
\[
h^{\mathrm{patch}}_{i,\ell,r} \leftarrow h^{\mathrm{clean}}_{i,\ell,r}.
\]
All other token positions and layers remain from the corrupted run. Token regions correspond to \stage{<graph>}, \stage{<patterns>}, \stage{<synthesis>}, and answer-start spans.


The patched run produces $y_{i,\ell,r}^{\mathrm{patch}}$. Recovery is
\[
R_{i,\ell,r}
=
\frac{M(y_{i,\ell,r}^{\mathrm{patch}})-M(y_i^{\mathrm{corr}})}
{M(y_i^{\mathrm{clean}})-M(y_i^{\mathrm{corr}})+\epsilon}.
\]
Values near zero indicate little recovery, values near one indicate recovery close to the clean run. We do not clip negative values.

We separately test the implementation with a strict identity control, in which the same cached activation is written back at the same layer and token position under deterministic generation; the results are reported in Table~\ref{tab:identity_control}.

\begin{tcolorbox}[
  enhanced jigsaw,
  breakable,
  title={\textbf{Graph-to-Answer Circuit Tracing} \hfill
  \textit{Question type: cross-domain mapping}},
  colback=stageblueback,
  colframe=stageblue,
  boxrule=0.7pt,
  arc=2mm,
  left=5pt,
  right=5pt,
  top=5pt,
  bottom=5pt,
  before skip=0.6em,
  after skip=0.6em,
  fonttitle=\small,
  title after break={\textbf{Graph-to-Answer Circuit Tracing}
  \hfill\textit{continued}},
  pad at break*=1mm
]

\smallskip
\noindent\textbf{Question (abridged).}
\textit{Biological immune systems and multi-agent AI frameworks both address open-ended problems through division of labor. Analyze where the analogy breaks down mechanistically, then propose one concrete capability next-generation multi-agent scientific systems would need to achieve adaptive, long-horizon robustness.}

\smallskip
\noindent\rule{\linewidth}{0.3pt}

\smallskip
\noindent\textbf{\stage{<graph>} stage (causal structure).}
\textit{Agent Experience $\to$ Dynamic Memory Encoding (hierarchical consolidation),
Dynamic Memory Encoding $\to$ Adaptive Strategy Refinement (feedback-driven learning),
Adaptive Strategy Refinement $\to$ Long-horizon Robustness $\uparrow$.}

\smallskip
\noindent\textbf{\stage{<synthesis>} stage.}
\textit{Next-generation multi-agent systems can achieve adaptive, long-horizon robustness by adopting a dynamic memory encoding and retrieval mechanism analogous to immune memory formation. Through feedback-driven updates and distributed knowledge sharing, the system evolves its response strategies over time.}

\smallskip
\noindent\rule{\linewidth}{0.3pt}

\smallskip
\noindent\textbf{Final answer (clean run).}
\textit{The proposed capability is a decentralized, hierarchical memory architecture that allows agents to store experiences, extract causal patterns, and iteratively refine strategies based on task feedback --- directly mirroring B-cell memory consolidation after antigen exposure.}

\smallskip
\noindent\rule{\linewidth}{0.3pt}

\smallskip
\noindent\textbf{Corrupted run} (wrong-graph corruption: graph replaced with an unrelated question's graph).
\textit{The proposed capability is a hierarchical, distributed communication protocol\ldots}
The answer mentions communication structure but loses the memory-encoding mechanism chain.

\smallskip
\noindent\textbf{Mechanism-F1 recovery at layer 36, synthesis-stage patch: $R = 0.70$.}
Under this corruption and recovery metric, the layer-36 synthesis intervention restores much of the measured mechanism content. This result is consistent with localized recoverability in the \stage{<synthesis>}-to-answer pathway, it does not establish that this is the model's only or complete causal route.

\end{tcolorbox}

This intervention-based analysis produces a recovery map showing where mechanism-related information is most recoverable under the specified corruption and metric.
\section{Results}

\subsection{Embedding similarity hides mechanism damage}
\label{sec:embedding_hides}

Graph corruptions often leave the final answer close to the clean answer in the embedding space. However, mechanism-F1 and graph-edge support drop under the same corruptions. This shows why a visualization based only on final-answer semantic similarity is misleading: the answer can remain fluent while the causal mechanism path collapses. \Cref{fig:edge_corruption} and the corruption-strength panel provide the sanity check that these graph perturbations break directed graph structure rather than simply rephrasing it.

\subsection{Full layer-by-region recovery landscape}
\label{sec:full_sweep}

\begin{figure*}[t]
\centering

\begin{minipage}{\linewidth}
\centering
\safeincludegraphics[width=\linewidth]{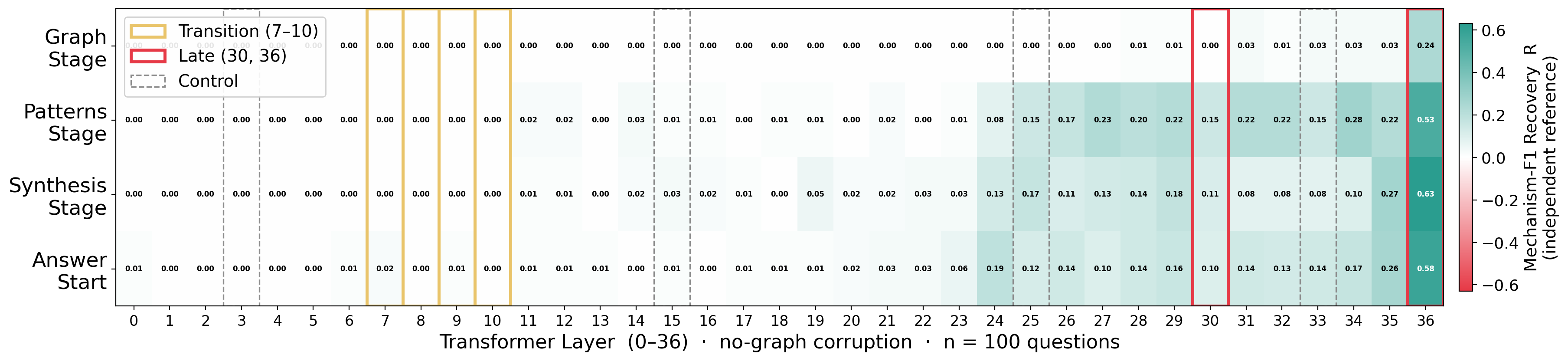}
\vspace{-0.8em}
\centerline{\small (a) No-graph corruption}
\end{minipage}

\vspace{0.6em}

\begin{minipage}{\linewidth}
\centering
\safeincludegraphics[width=\linewidth]{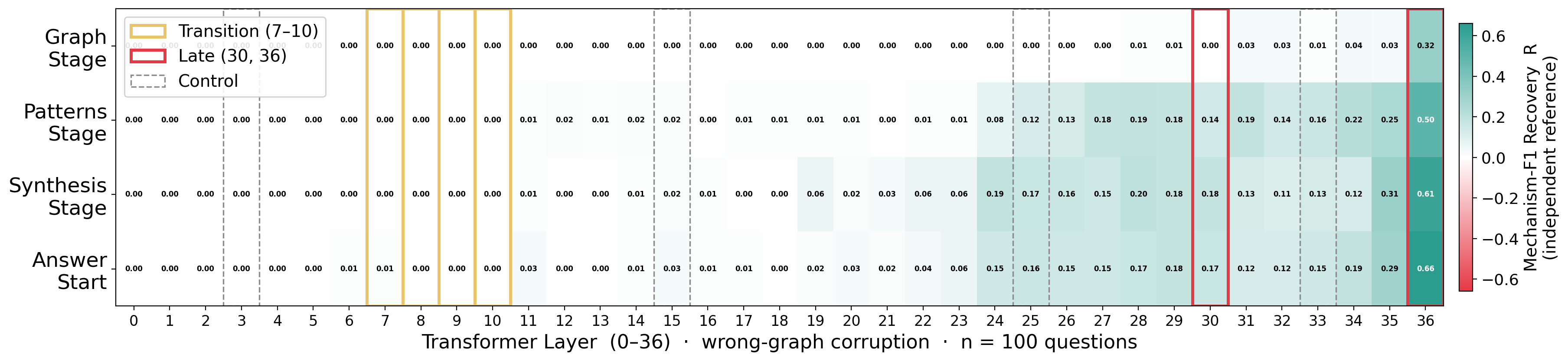}
\vspace{-0.8em}
\centerline{\small (b) Wrong-graph corruption}
\end{minipage}

\vspace{-0.4em}
\caption{Full layer-by-token-region recovery grids over all 100 questions. 
Orange outlines mark transition layers 7-10, red outlines mark late layers 30 and 36, dashed gray outlines mark control layers. 
(a) Under no-graph corruption, recovery is concentrated in late synthesis and answer-start regions, especially layer 36. 
(b) Under wrong-graph corruption, the same pattern holds even when the model receives graph-like text from the wrong question. 
Across both settings, layers 7--10 show little mechanism recovery. To read the grid, compare token-region columns within a row, then compare early, control, and late rows across corruption panels. The persistent late synthesis/answer-start concentration is the main visual pattern.}
\label{fig:full_sweep}
\end{figure*}


\Cref{fig:full_sweep} shows the full layer-by-token-region mechanism recovery grids for no-graph and wrong-graph corruptions over all 100 questions. The shuffled-graph grid shows the same qualitative pattern and is included in the supplemental material. The orange region marks layers 7-10, which were identified by a reasoning-answer divergence bump in an earlier representation analysis. The red region marks late layers 30 and 36. The full grid shows that the transition region is not where mechanism content is recovered. Recovery remains near zero across layers 7-10. In contrast, recovery increases in late layers, especially for synthesis-stage and answer-start tokens. The shuffled-graph grid shows the same qualitative pattern.

This result changes our interpretation of layers 7-10. They are not a storage location for the final answer mechanism. Rather, they appear to be a representational transition region where reasoning and answer states begin to separate. The recoverable mechanism appears later, after the graph structure has been transformed into synthesis and answer-start representations.


\subsection{Late layers outperform control layers}
\label{sec:late_layers}

Table \ref{tab:full_sweep_bootstrap} summarizes the full-sweep result with paired bootstrap resampling over all 100 questions. Late layers $\{30,36\}$ recover more mechanism-F1 than control layers for all three corruptions: no graph ($\Delta=+0.237$), wrong graph ($\Delta=+0.262$), and shuffled graph ($\Delta=+0.224$). The corresponding transition-vs-control comparisons are not positive, supporting the distinction between transition and recovery.

\begin{table}[t]
\centering
\small
\caption{Full-sweep bootstrap summary over all 100 questions. Late layers are \(\{30,36\}\), control layers are \(\{3,15,25,33\}\).}
\label{tab:full_sweep_bootstrap}
\begin{tabular}{lccc}
\toprule
\textbf{Corruption} & \textbf{Late--control} & \textbf{95\% CI} & \textbf{Result} \\
\midrule
No graph & $+0.237$ & $[+0.208,+0.265]$ & $p<0.001$ \\
Wrong graph & $+0.262$ & $[+0.229,+0.294]$ & $p<0.001$ \\
Shuffled graph & $+0.224$ & $[+0.197,+0.252]$ & $p<0.001$ \\
\bottomrule
\end{tabular}
\end{table}

\subsection{Negative controls, identity check, and extractor validation}
\label{sec:controls_validation}

\begin{figure}[t]
\centering
\safeincludegraphics[width=\columnwidth]{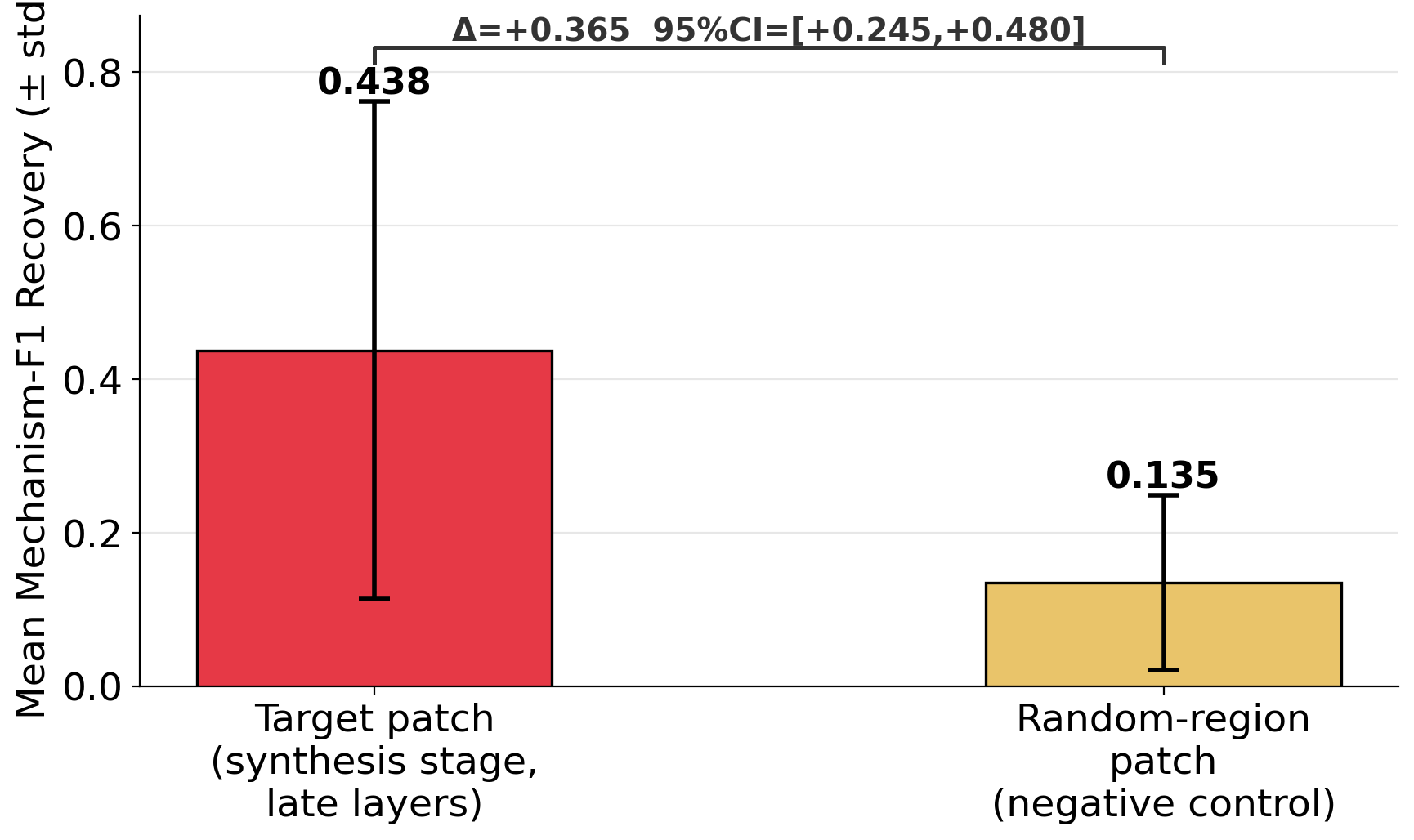}
\caption{Negative control for patching specificity. Targeted late-layer synthesis patching recovers more mechanism-F1 than random-region patching under wrong-graph corruption ($\Delta=+0.365$, 95\% CI $[+0.245,+0.475]$, $p<0.001$).}
\label{fig:negative_control}
\end{figure}

\Cref{fig:negative_control} shows a negative control. If recovery were caused by injecting arbitrary clean activations, random-region patching would recover similar mechanism content. Instead, targeted late-layer synthesis patching recovers substantially more mechanism-F1 than a random-region control. This supports the claim that recovery is localized to the graph-to-synthesis-to-answer pathway rather than being a generic clean-activation effect.

\begin{table}[t]
\centering
\scriptsize
\caption{Implementation checks for deterministic inference and strict identity
patching on the diagnostic example. The identity patch writes the same cached
activation back at the same layer and token position.}
\label{tab:identity_control}
\begin{tabular}{p{0.25\columnwidth}p{0.27\columnwidth}p{0.34\columnwidth}}
\toprule
\textbf{Check} & \textbf{Compared output} & \textbf{Result} \\
\midrule
Fixed-token repeat
& All token logits
& Exact match; maximum absolute difference $=0$ \\
\addlinespace
Deterministic generation
& Token sequence and decoded text
& Both identical \\
\addlinespace
Strict identity patch
& Logits, sequence, and decoded text
& All identical; maximum absolute logit difference $=0$ \\
\bottomrule
\end{tabular}
\end{table}

To check that the recovery results were not produced by the intervention code, we ran three implementation checks (Table~\ref{tab:identity_control}). Repeating the fixed-token forward pass produced exactly the same logits. Two greedy generations also produced the
same token sequence and decoded text. We then cached the residual-stream activation at the tested layer and token position and wrote that same tensor back during an otherwise identical generation. No logit entries changed, the maximum absolute logit difference was zero, and the generated sequence and text remained unchanged. The patching operation is therefore neutral when
the activation, its position, and the generation trajectory are held fixed. This result also changes how we interpret the earlier clean-run sensitivity check. The broader same-question condition changed mechanism-F1 by $-0.234$ on average, compared with $-0.379$ for the random-question condition. These values are not strict identity-patch effects: the same-tensor identity replacement leaves the output unchanged. The nonzero changes must therefore
come from another part of the broader intervention or scoring procedure, rather than from the activation overwrite alone. We retain these results as sensitivity controls, but do not use them as evidence that the patching hook itself damages the model's output.

\begin{figure}[t]
\centering
\safeincludegraphics[width=\columnwidth]{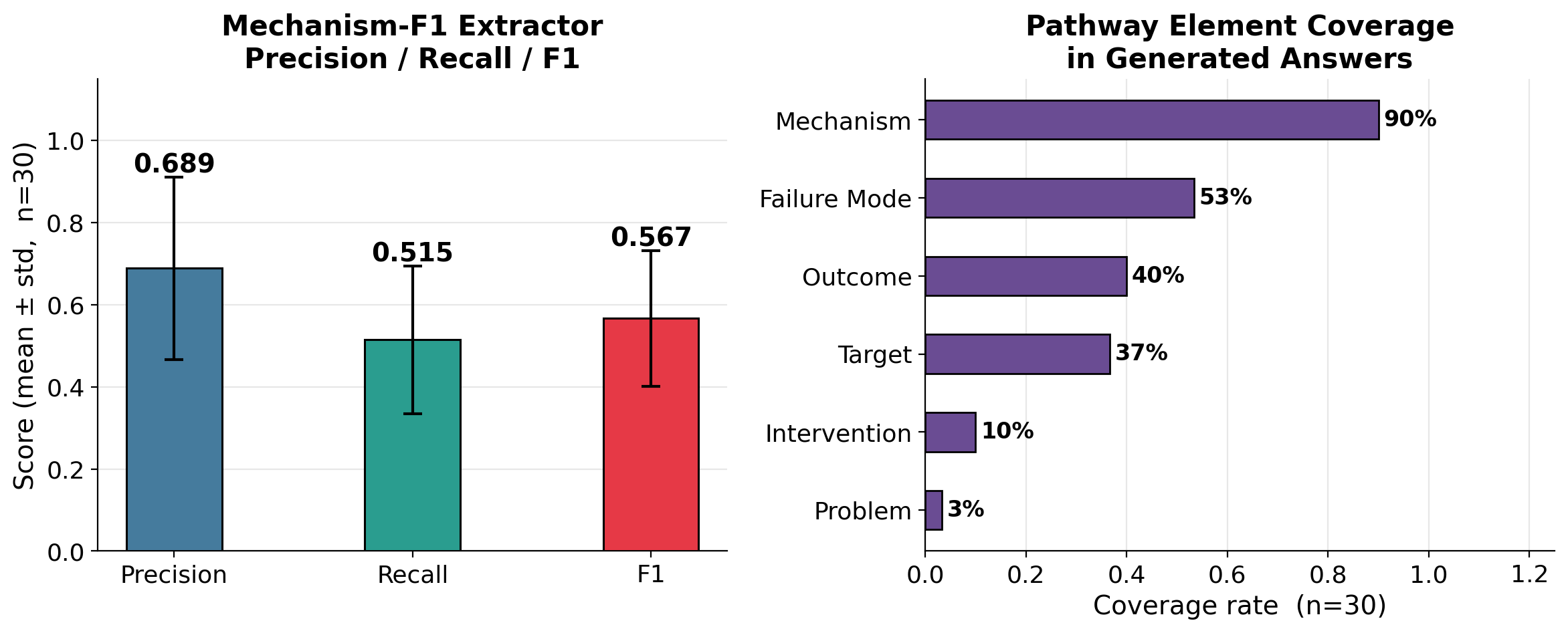}
\caption{Mechanism-F1 extractor validation. A 30-example audit gives precision 0.689, recall 0.515, and F1 0.567. The extractor most reliably captures mechanism terms, so we use mechanism-F1 as a relative recovery proxy rather than an absolute correctness measure.}
\label{fig:extractor}
\end{figure}

Finally, \Cref{fig:extractor} reports a manual audit of 30 generated answers. The extractor obtains a precision of 0.689, a recall of 0.515, and an F1 score of 0.567. It captures central mechanism terms more reliably than fields such as the problem context and intervention, and
it can miss relations expressed through paraphrases. We therefore use mechanism-F1 only for relative comparisons among matched clean, corrupted, and patched runs, rather than as an absolute measure of scientific correctness.

We also performed a trace-faithfulness audit, where we evaluated 15 examples with a structured reviewer sheet that asked whether the final answer mechanism was supported by the visible \stage{<graph>} and \stage{<synthesis>} stages. This audit was done using Claude Fable 5 as the judge. It is used as a qualitative check on the automatic mechanism-F1 score. The audit marked 15/15 examples as supported and 15/15 as faithful, but 5/15 as partial. The partial cases reveal two recurring issues: an underspecified mediator and solution steps stated in synthesis but not fully derived from the graph. The audit also identified false negatives in automatic mechanism-F1 caused by lexical mismatch between equivalent causal connectives. We therefore report mechanism-F1 as a relative recovery proxy, not as an absolute measure of trace faithfulness.

\begin{table}[t]
\centering
\small
\caption{Trace-faithfulness audit over 15 examples. The audit is
used as a qualitative check using Claude Fable 5.}
\label{tab:faithfulness_audit}
\begin{tabular}{lc}
\toprule
\textbf{Audit item} & \textbf{Count} \\
\midrule
Answer mechanism supported by graph/synthesis & 15/15 \\
Answer causal chain faithful to visible reasoning & 15/15 \\
Partial / underspecified trace & 5/15 \\
Fully supported without partial flag & 10/15 \\
\bottomrule
\end{tabular}
\end{table}


\section{Discussion}
\label{sec:discussion}

The main finding is a separation between representational transition and measured mechanism recovery. Layers 7--10 were previously highlighted because reasoning and answer representations begin to diverge there, but the full recovery grid shows little recovery in that region. Recovery instead concentrates in late synthesis and answer-start regions. A single selected-layer plot would make this distinction easy to miss, the full grid exposes both the negative result in the transition region and the repeated late-layer pattern across corruptions.

\paragraph{How the visual workflow supports auditing.}
The intended workflow begins with an answer that may appear plausible. A scientist first checks whether graph corruption damages its directed mechanism while leaving surface similarity high. The recovery grid then answers three diagnostic questions: which trace stage carries the strongest recoverable signal, whether that signal is localized or diffuse across layers, and whether it persists across different corruptions. The user can follow a suspicious or high-recovery cell back to the corresponding graph, synthesis, and generated answer. Aggregate scores cannot support this case-level navigation or reveal whether a mean effect is driven by one narrow region. The current paper demonstrates these analytical views statically, an interactive system with filtering, linked selection, and side-by-side trace comparison is a natural next step.

\paragraph{Implications for AutoSci.}
In an automated-science pipeline, this workflow is most useful as a human-controlled gate between hypothesis generation and costly downstream action. A fluent hypothesis whose mechanism collapses under graph perturbation can be regenerated or routed to expert review. A stable, localized pattern can guide model debugging, targeted data collection, or the placement of monitoring probes. The present results do not automatically repair a hypothesis or decide whether it is scientifically true. They narrow the search: they show where the model's graph-to-answer pathway deserves inspection and provide evidence that answer-level similarity alone is insufficient.

The relevant questions are therefore not only ``Does the answer sound right?'' but also ``Does its mechanism survive perturbation, where is that mechanism recoverable, and is it supported by the visible trace?''
\section{Limitations}

This is a single-model, single-domain case study over 100 questions. The results should not be generalized to other model families, scientific domains, larger benchmarks, or closed models without replication. The benchmark is intentionally open-ended and mechanism-focused rather than a standard field benchmark, and its construction may shape the observed effects.

Mechanism-F1 is a relative proxy, not a measure of experimental truth. The automatic extractor has precision 0.689, recall 0.515, and F1 0.567, so missed paraphrases or relations can alter recovery estimates. The 15-example trace audit is preliminary and model-judged rather than an independent materials-expert study. A stronger validation would use blinded human comparisons of late- versus transition-layer outputs and an alternative relation score.

The intervention result is conditional on the corruption, residual-stream location, token alignment, decoding procedure, and normalization. A strict identity patch was neutral: writing the same cached activation back at the same layer and token position left the logits, generated sequence, and text unchanged. Broader same-question and random-question sensitivity checks still
produced nonzero changes, showing that transfers between different computational trajectories can affect the output. We therefore claim localized recoverability under a specified intervention and metric, not the discovery of a complete head-, neuron-, or circuit-level mechanism. Exact token alignment remains important because removing or replacing the graph
can change the length and position of later regions.

Finally, the current visual contribution consists of static diagnostic views. We have not evaluated an interactive interface or measured whether these views improve scientists' decisions. A future study should implement linked trace/heatmap inspection and evaluate whether domain experts detect unsupported hypotheses more accurately or efficiently.
\section{Conclusion}

We presented a visual diagnostic workflow for tracing graph-to-answer mechanism recovery in materials-science hypothesis generation. The workflow links readable reasoning stages, controlled graph corruptions, mechanism-sensitive scoring, and full layer-by-token-region recovery grids. In this Graph-PRefLexOR-8B case study, layers 7--10 behave as a representational transition zone, whereas the strongest measured recovery appears later in synthesis and answer-start regions. The practical implication for auditable AI co-scientists is not that a heatmap proves scientific correctness, but that it provides a human checkpoint: plausible answers can be tested for mechanism damage, localized for closer inspection, and routed for regeneration or expert review before downstream action.


\acknowledgments{This work was supported by the U.S. Department of Energy, Office of Science, Office of Advanced Scientific Computing Research and Office of Basic Energy Sciences, Scientific Discovery through Advanced Computing (SciDAC) program under the FORUM-AI project. This work was primarily performed at the Oak Ridge National Laboratory, which is managed by UT-Battelle, LLC, for the U.S. DOE under the contract No. DE-AC05-00OR22725.}

\section{Use of Large Language Models}
Large language models (LLMs) were used for evaluation and question refinement. All LLM-generated materials and outputs were reviewed, filtered, and analyzed by the authors.

\bibliographystyle{abbrv-doi}
\bibliography{template}
\balance
\end{document}